\documentclass[journal]{IEEEtran}
\usepackage{cite}

\ifCLASSINFOpdf
\else
\fi
\usepackage[]{graphicx}
\usepackage{hyperref}
\usepackage{ccaption}

\usepackage[cmex10]{amsmath}
\usepackage{amssymb}
\usepackage{url}

\usepackage{epstopdf}
\usepackage{multirow}
\usepackage{subfigure}
\usepackage[subnum]{cases}

\onecolumn

\begin{document}

%
\title{Automatic Data Registration of Geostationary Payloads for Meteorological Applications at ISRO}
%
%
\author{Jignesh S. Bhatt and N. Padmanabhan

\thanks{Corresponding author: Jignesh S. Bhatt (email: jignesh.bhatt@iiitvadodara.ac.in) is with Indian Institute of Information Technology (IIIT), Vadodara, Building \#9, Government Engineering College, Sector - 28, Gandhinagar 382028, India.} 

}

\maketitle

\begin{abstract}
The launch of KALPANA-1 satellite in the year 2002 heralded the establishment of an indigenous operational payload for meteorological predictions. This was further enhanced in the year 2003 with the launching of INSAT-3A satellite. The software for generating products from the data of these two satellites was taken up subsequently in the year 2004 and the same was installed at the Indian Meteorological Department, New Delhi in January 2006. The VHRR sensor onboard KALPANA-1 operates in the visible (0.55 to 0.75 $\mu$m), water-vapour (5.7 to 7.1 $\mu$m) and thermal infrared (10.5 to 12.5 $\mu$m) wavelength bands with 2 Km and 8 Km spatial resolutions; and the CCD sensor onboard INSAT-3A covers the visible (0.62 to 0.68 $\mu$m), near-infrared (0.77 to 0.86 $\mu$m), and the short-wave infrared (1.55 to 1.69 $\mu$m) bands with 1 Km ground resolution. Registration has been one of the most fundamental operations to generate almost all the data products from the remotely sensed data. Registration is a challenging task due to inevitable radiometric and geometric distortions during the acquisition process. Besides the presence of clouds makes the problem more complicated. In this paper, we present an algorithm for multitemporal and multiband registration. In addition, India facing reference boundaries for the CCD data of INSAT-3A have also been generated. The complete implementation is made up of the following steps: 1) automatic identification of the ground control points (GCPs) in the sensed data, 2) finding the optimal transformation model based on the match-points, and 3) resampling the transformed imagery to the reference coordinates. The proposed algorithm is demonstrated using the real datasets from KALPANA-1 and INSAT-3A. Both KALAPANA-1 and INSAT-3A have recently been decommissioned due to lack of fuel, however, the experience gained from them have given rise to a series of meteorological satellites and associated software; like INSAT-3D series which give continuous weather forecasting for the country. This paper is not so much focused on the theory (widely available in the literature) but concentrates on the implementation of operational software.   
\end{abstract}  


\IEEEpeerreviewmaketitle

\section*{Nomenclature}
\begin{center}
ISRO Indian Space Research Organization;\\
SAC Space Applications Centre;\\
IMD Indian Meteorological Department;\\
DPS Data processing system;\\
DPSG Data Products Software Group;\\
PSLV Polar satellite launch vehicle;\\
GEO Geostationary Orbit;\\
CCD Charge coupled device;\\
VHRR Very high resolution radiometer;\\
OLR Outgoing long-wave radiation;\\
QPE Quantitative precipitation estimation;\\
UTH Upper troposheric humidity;\\
AMV Atmospheric motion vectors;\\
NDVI Normalized difference vegetation index;\\
MOSDAC Meteorological \& Oceanographic Satellite Data Archival Centre;\\
UT Universal time.\\
\end{center}

\section{Introduction}
\IEEEPARstart{I}{ndia} has a long history of space research activities. This was started in 1963 with the launch of a two stage rocket.  Dr Vikram Sarabhai\footnote{\url{www.vssc.gov.in/internet/}}, the father of India's space program, had initiated these activities in the pursuit of using space research for peaceful purposes leading to the betterment in the life of the common man.
												         
Indian space research organization (ISRO) is aimed to harness space technology for national development\footnote{\url{https://www.isro.gov.in/}}. One of the primary centers of ISRO is the Space Applications Centre (SAC)\footnote{\url{www.sac.gov.in/}}. The basic objectives of SAC are design and development of optical and microwave remote sensing payloads, communication transponders, software for deriving data products from satellite imagery, and other application related activities. To this end, SAC has two broad areas of operation, viz., satellite based communication, including television and remote sensing for natural resources survey and management as well as meteorology.

KALPANA-1 and INSAT-3A were GEO satellites to study the weather patterns as well as the ground vegetation index. KALPANA-1 is having VHRR payload, launched by a PSLV into the geostationary orbit by reducing the weight of the payload. INSAT-3A is basically a communication satellite having CCD and VHRR imaging instruments. Incoming images from these satellites are in multiple bands, acquired at different times in a day. The image data are processed so as to generate different levels of products related to atmospheric predictions and vegetation index generation. For instance, OLR, QPE, UTH, and AMV are major geophysical products from the VHRR data and NDVI from the CCD data. 

In order to generate such products accurately, the images should be {\itshape registered}. Image registration is defined as the process of overlaying two images of the same scene taken at different times, from different viewpoints, and/or by different sensors , i.e., the registration of two images is the process by which they are aligned so that overlapping pixels correspond to the same entity that has been imaged\cite{Brownvol.24pp.326-3761992, Zitovavol.21pp.977-10002003., Nag2017}. Any two pixels at the same location in both images are then `in register" and represent two samples at the same point on the earth \cite{Schowengerdt2007}.  
\begin{figure*}[th!]
\hspace{0.8in}
\includegraphics[width=1.0\textwidth]{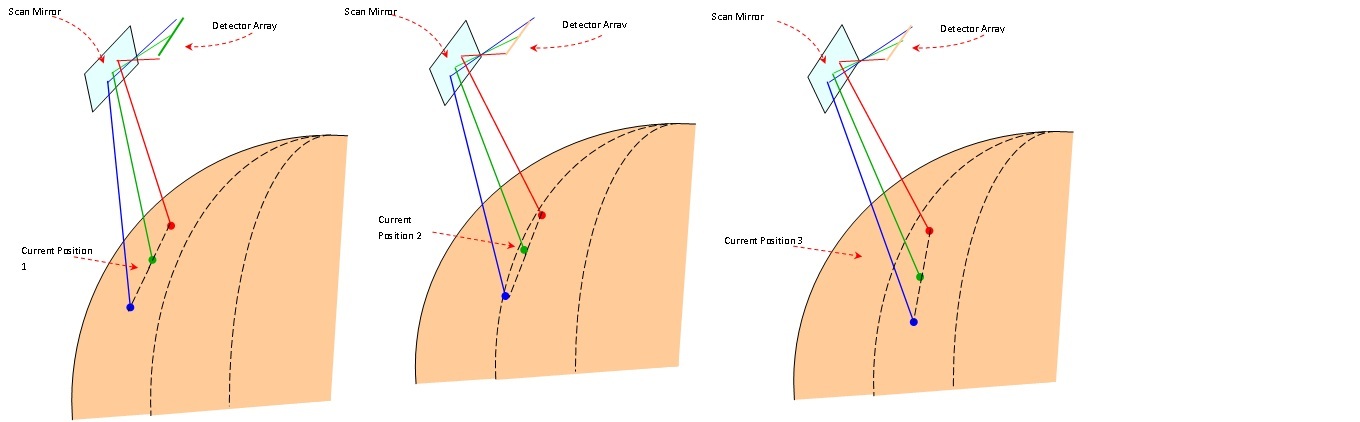}
\caption{GEO imaging using the scanning mirror mechanism. It scans in-line from west to east, then down a line to again scans in-line, and so on.}
\end{figure*}
The ideal image is an accurate reproduction of ground reflectance in the various spectral bands. However, most recorded images suffer from various degrees of distortions. There are two ways in which satellite images can be distorted, viz. radiometric distortions and geometric distortions. The major causes of the radiometric distortions are due to the atmosphere and imaging systems errors, while the panoramic distortions induced by perspective projections due to the earth curvature are the main reasons behind the geometric distortions in the sensed imagery \cite{Padmanabhanvol.86no.8pp.1113-11212004.}. Besides, there are many random distortions mainly due to unpredictable occurrences such as variations in platform positions, altitude variations, etc. thus leading to unpredictable non-linear distortions across the image area \cite{Padmanabhanvol.86no.8pp.1113-11212004.}. Hence, there is continuous need for the correcting algorithms that register the data to the reference coordinates. This in turn enables the accurate generation of almost all the data products in the remote sensing. In this paper, we present multitemporal and multiband registration of image data for both the meteorological payloads, i.e., KALPANA-1 and INSAT-3A. India facing reference boundaries for the CCD imagery from INSAT-3A have been derived and overlaid on the registered data. This further validates the accuracies of the registration algorithm. 

The rest of the paper is organized as follows. Section II gives overview of the imaging from GEO including the sensors details and the data processing system deployed at the SAC (ISRO). In section III we provide real examples detailing the need for registration in the remote sensing data. The complete proposed approach for performing the registration is dealt in section IV. Section V demonstrates the performance with the series of experiments conducted on the real data collected by KALPANA-1 and INSAT-3A. Finally, section VI concludes the paper with the key achievements.

\section{Overview of GEO Imaging and Data Processing System}
The geostationary orbit is used to get the full coverage of the globe for the meteorological purposes. The image data from two INSAT, viz., KALPANA-1 and INSAT-3A, were being used for the meteorological related studies. KALPANA-1 had VHRR payload and INSAT-3A had got CCD based imager and VHRR sensor as well apart from the communication transponders. Fig. 1 provides schematic diagram of the scanning mirror mechanism that remotely acquired the data using the GEO platform. More details are available at MOSDAC\footnote{\url{https://mosdac.gov.in/content/Mission/kalpana-1/payloads}}.
\begin{figure*}[th]
	\centering
		\includegraphics[width=0.8\textwidth]{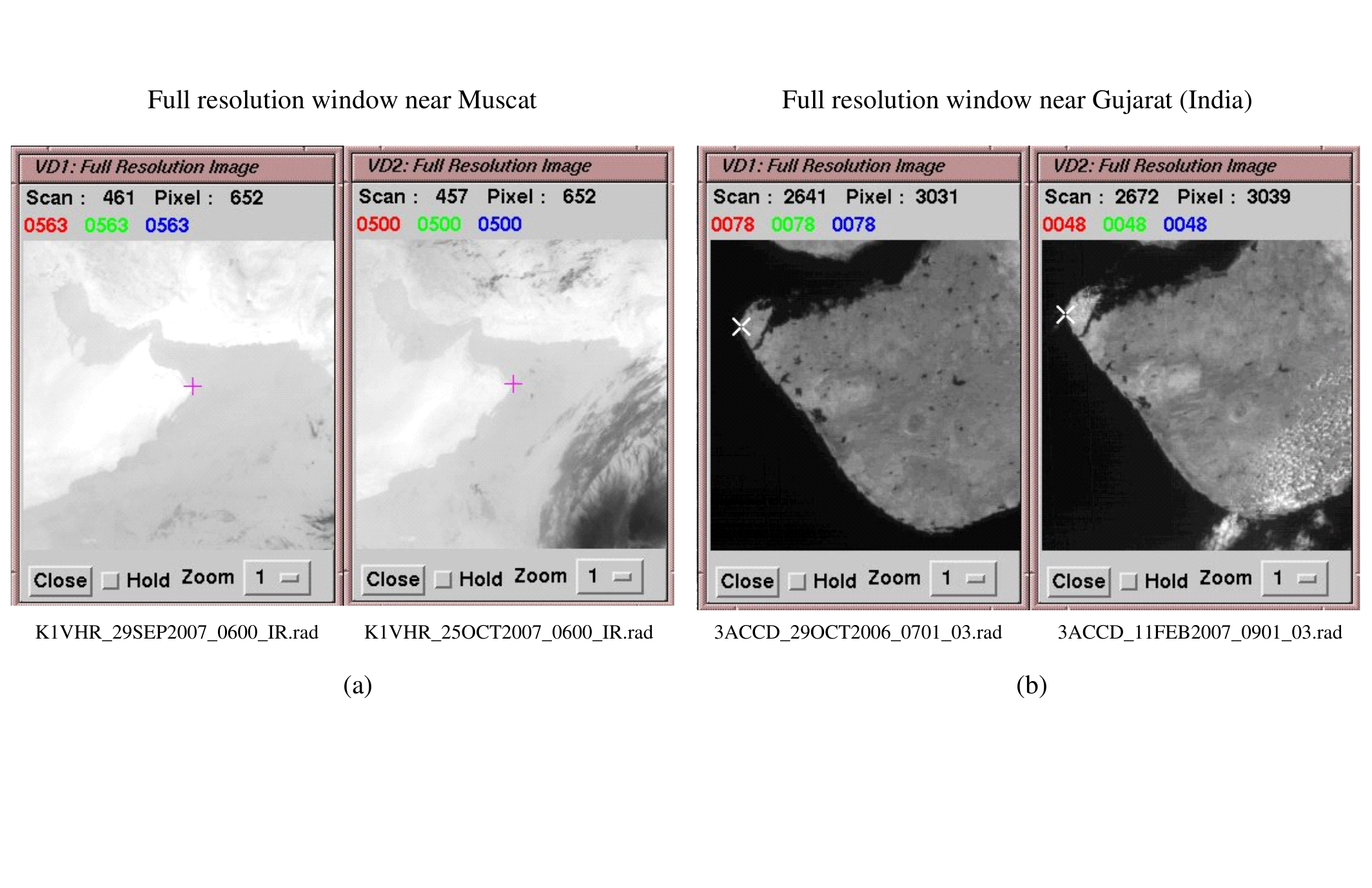}
	\caption{Need for image registration in the VHRR and CCD data acquired on two different days. (a) shift in VHRR images near Muscat, and (b) shift in CCD images near Gujarat (India). The images are displayed using the SAC Image Display Software.}
	\label{fig:need_CCD}
\end{figure*}
\subsection{VHRR Payload}
KALPANA-1 and INSAT - 3A are equipped with the VHRR imaging instrument and the scanning mirror mechanism (see Fig. 1). It scans in-line from west to east, then down a line and again scans in-line from east to west, and so on. Imaging is done once an hour, and a complete scan takes about 33 minutes. Thus, a total of 24 image frames are available in a day. 
\subsection{CCD imager}
INSAT-3A payload has a basic telescope and scan mechanism for imaging as shown in Fig. 1. Two diachronic beam splitters achieve the separation of the three bands. The scan mirror is mounted on a two-axis gimbaled scan mechanism system to generate a two-dimensional image by sweeping the detector array field-of-view across the earth's surface in west-east (fast-scan) and north-south (slow-scan) directions. The long axis of detector arrays is oriented in the north-south direction. The detector array contains 300 detector elements. The fast-scan line is scanned from west to east by the scan mirror in approximately one minute. At the end of the line, the mirror retraces to the west, is stepped south by $0.3955^{\circ}$ and starts imaging the next line. Three hundred contiguous image lines are generated for each of the bands during one fast-scan line, and this is called a frame. The step down between two successive fast-scans is done such that there is an overlap of about 17 lines between the two successive frames. Each frame covers a field-of-view of $10^{\circ}$ in the west-east direction and about $0.3985^{\circ}$ in the north-south direction. Hence, resolution will be varying from one edge to another because of panoramic distortion. Four images from each band are acquired in a day, at 0300 UT, 0500 UT, 0700 UT and 0900 UT.
\subsection{Data reception and processing system at SAC (ISRO)}
The sensed data is received by the earth station and in turn processed by the image data processing system. The meteorological image data (MET DATA) from both CCD and VHRR sensors are digitally stored in a data acquisition system. The data is then coupled with a real-time data reception system through a computer interface unit. In turn, the header and framing information is added on the data. The images are then available for geometric and radiometric corrections at data processing system (DPS). Only after successful data registration, the images are available for the generation of various data products by applying appropriate algorithms.  

\section{Examples of Need for Registration in KALPANA-1 and INSAT-3A Data}
In this section, we illustrate the need for image registration in the VHRR and CCD data using a real example for each. A scenario is discussed in Fig. 2 using two ground control points (GCPs). It should be noted that the images are displayed using the SAC Image Display Software developed at DPSG. Fig. 2 (a) points out a need for registration in a VHRR image acquired by the KALPANA-1 payload. The same portion of two images of the same (IR) band and same time (0600), but acquired on different dates are shown in full resolution windows. As shown in the Fig. 2 (a), a point near Muscat (i.e., scan: 461, pixel: 652) in the first image, is shifted to scan: 457, pixel: 652 in the second image. Similarly, Fig. 2 (b) depicts the need for data registration in a CCD image acquired by the INSAT-3A imager. The same portion of two images of the same band acquired on different dates and times are shown. A same point on Gujarat state (India) (i.e., scan: 2641, pixel: 3031) in the first image, is shifted to scan: 2672, pixel: 3039 in the second image. Such random deviations over the sensed area raise the need for the image registration in meteorological imagery. Considering the ground (spatial) resolution of both the imagery, it is clear that these data need to be registered with their respective GCPs, and hence required to be corrected for generation of accurate data products for further analysis.

\section{Proposed Approach for Automatic Data Registration}
In this section, we discuss our algorithm for registering the meteorological data. Registration of images (sensed images) is to be done with a reference image which is considered as a template. First the problem is defined along with block diagram of a general registration process. We describe the operational blocks with the theoretical details. Toward the end of this section, we present a simplified flowchart of the software implementation.  
\begin{figure}[h]
	\centering
		\includegraphics[width=0.5\textwidth]{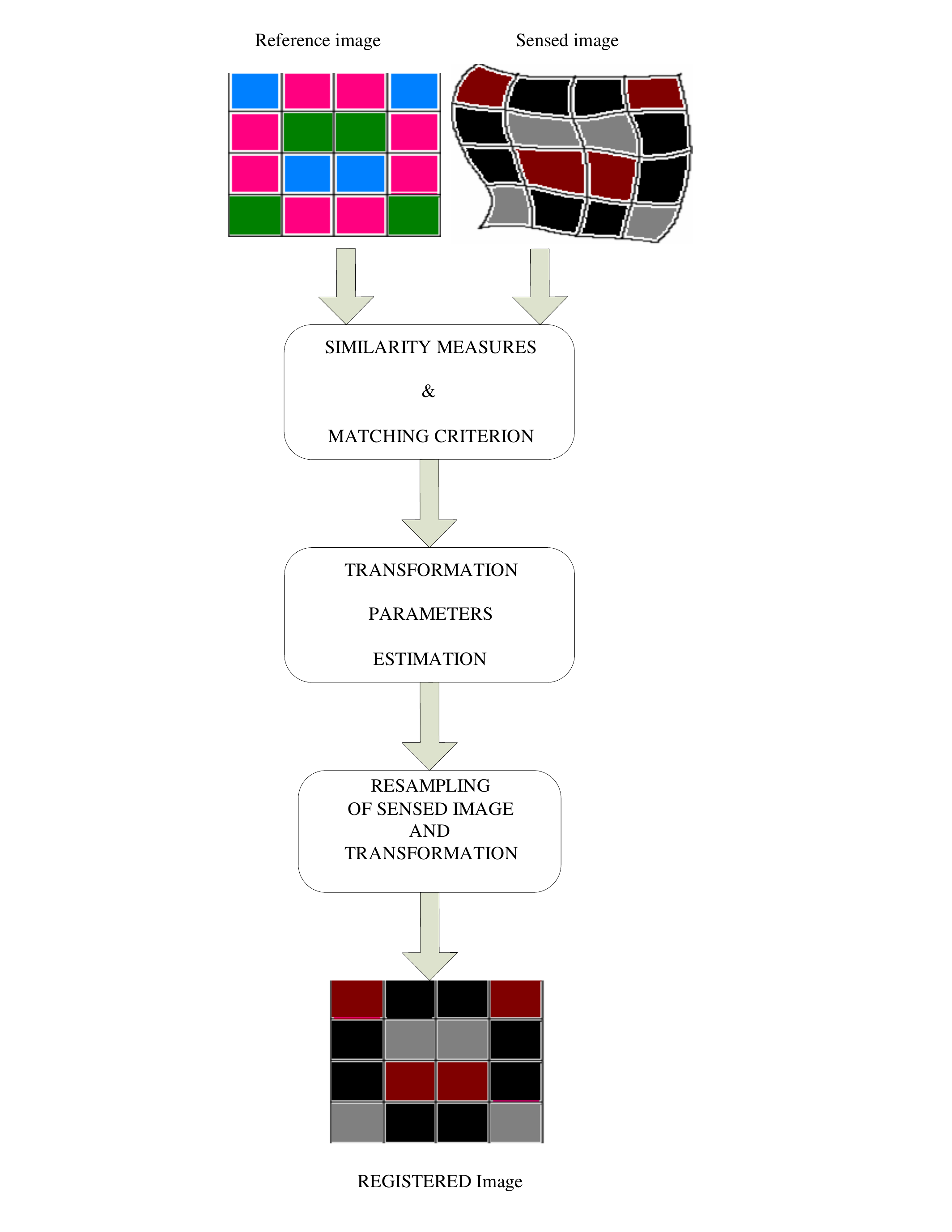}
	\caption{Approach for data registration of KALPANA-1 and INSAT-3A.}
	\label{fig:proposed_approach}
\end{figure}

As shown in a general schematic Fig. 3, the incoming images of the KALPANA -1 and INSAT - 3A are treated as the sensed images and hence required to be corrected geometrically with respect to the reference template while radiometric statistics of a registered image should be close (ideally same) to the sensed image. A set of ground control points (GCPs) may be selected or given from the reference image. The algorithm should automatically be able to identify the corresponding match points in the sensed image, find the most appropriate transformation from that set of match points, and perform the necessary resampling in order to warp the sensed image onto the reference coordinates. The algorithm should be robust enough to avoid any false point matching and consequent misregistration. The software should be generalized enough for possible future modifications, whenever necessary. Importantly, the complete process should take minimum amount of time to execute. 

\begin{figure}[h]
	\centering
		\includegraphics[width=0.5\textwidth]{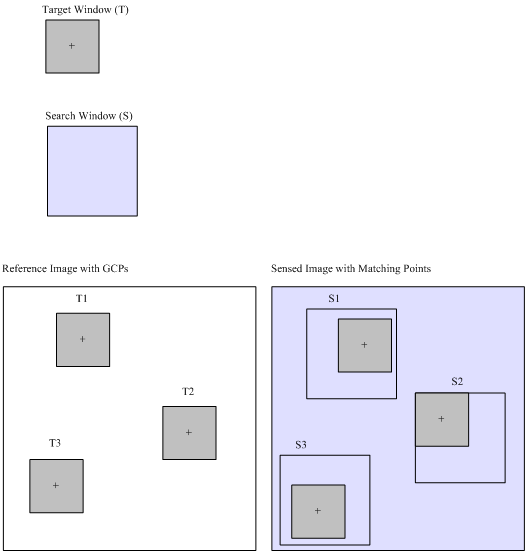}
	\caption{Moving target window algorithm to automate the match-point identification process in the sensed image.}
	\label{fig:proposed_approach}
\end{figure}

{\itshape Similarity measures and matching criterion:} As mentioned in the problem definition, a sensed image is to be registered with a reference image. Hence, the first step in image registration is identifying corresponding features between the reference and a sensed images. There would be, in general, non-linear distortions present in the incoming satellite imagery. We implemented moving window algorithm to automate the match-point identification process as follows. Referring to Fig. 4, a small target window (T) is taken at a GCP in the reference image. A larger search window (S) in the sensed image is considered at the same location, i.e., coordinates of the GCP in reference image.  The target window is moved within the search window pixel-by-pixel and scan-by-scan, as shown schematically in the Fig. 4. For each combination of the two windows, similarity measure is calculated and stored. At the end of process, the best matching position of the target window is found out based on the matching criterion of similarity measure. The entire process is repeated for all GCPs.

The similarity measure is a key step in the process of remote sensing data registration. The adopted measure should be robust, accurate and objective. Apart from the two classical measures, namely, normalised cross-correlation coefficient (NCC) and minimum sum of squared difference (MSD); two measures based on statistical dependence, namely, cluster reward algorithm (CRA) \cite{1024955} and mutual information (MI)\cite{Chenvol.41no.11pp.2445-2454.2003, 5340570} have been tested for the data. 

We now define the similarity measures along with their matching criterion. The SSD is defined as,

\begin{equation}
SSD(i,j) = \sum_{i=1}^{N} \sum_{j=1}^{N} \left[r_{ij} - s_{ij}\right]^{2},
\end{equation}
where, $N$ is the total number of rows or columns in the image, $r_{ij}$ and $s_{ij}$ are the brightness values at reference and sensed images, respectively, while $i,j$ are the image coordinates. This measure estimates squared differences between pixel brightness values in a reference and a sensed image. The point at which the SSD is least gives the location of match point. Thus, it is also called minimum SSD (MSD).

The NCC is defined as\footnote{ 
\begin{equation}
NCC(i,j) = \frac{\sum_{m=1}^{N} \sum_{n=1}^{N} \left[r(m,n)-\mu_{r}\right] \left[s(i+m,j+n)-\mu_{s}\right]}{\sqrt{\sum_{m=1}^{N} \sum_{n=1}^{N} \left[r(m,n)-\mu_{r}\right]^{2} \sum_{m=1}^{N} \sum_{n=1}^{N} \left[s(i+m,j+n)-\mu_{s}\right]^{2}}},
\end{equation}
}
where, $\mu_{r}$ and $\mu_{s}$ are the means of target and search windows, respectively. It is the well-known cross-correlation coefficient whose value lies between $-1$ and $+1$. The point at which the NCC is maximum, gives the location of match point.  

Besides, we tested two statistical measures in order to find the match points in a sensed image. First is the cluster reward algorithm (CRA)\cite{1024955} and it is based on joint histogram of reference and sensed images. Let $H_{RS}(k,l)$ be the joint histogram of the reference and the sensed images, and $H_{R}(k)$ and $H_{S}(l)$ be the histograms of reference image and sensed image, respectively. The parameter $I_{CRA}$ will have a high value when the joint histogram has little dispersion and that is the point of best match between the two images. It is defined as,
\begin{equation}
I_{CRA} = \frac{\frac{\phi}{F}-\frac{F}{P^{2}}}{1-\frac{F}{P^{2}}},
\end{equation}
where, $\phi=\sum_{k=0}^{N-1} \sum_{l=0}^{N-1} H_{RS}^{2}(k,l)$; $F=\sqrt{h_{r}h_{s}}$, $h_{R} = \sum_{k=0}^{N-1} H_{R}^{2}(k)$, $h_{s} = \sum_{l=0}^{N-1} H_{S}^{2}(l)$, and $P=N\times N$. 

Finally we use the mutual information (MI) which is another measure of statistical dependence between two images. The entropy, a classical measure of disorder between two random variables, is the crux in finding MI. The joint entropy found out of joint histogram of reference and sensed images is the clue in MI measure. The point at which MI value is maximum, gives the location of best match between the images. The MI\cite{Chenvol.41no.11pp.2445-2454.2003, 5340570} between two images is defined as,
\begin{equation}
MI(r,s) = H(r) + H(s) - H(r,s),
\end{equation}
where, $H(r)$, $H(s)$, and $H(r,s)$ are entropy of reference and sensed images, and the joint entropy, respectively.

In this work, we have used combination of NCC and MSD as a similarity measure, i.e., we consider a match point when NCC is highest and MSD is minimum while running the moving window algorithm at the GCPs. This combined similarity matching criteria is found faster and accurate for both KALPANA-1 and INSAT-3A image data. Also, we have observed that the MI and CRA as similarity measures may lead to false point matching that ultimately can lead to misregistration. In addition, both the statistical measures are computationally taxing.  

{\itshape Warping:} After the match points identification, we aim at establishing a transformation that can correct the distortions in the sensed image. The idea is to reposition pixels from their original locations in the data array into a specified reference coordinates. There are three components to the process: (i) selection of suitable mathematical distortion model, (ii) coordinate transformation, and (iii)	resampling. These three components are collectively known as {\itshape warping}.

The selection of an appropriate transformation model is critical to the accuracy of any distortion correction. It is difficult to supply precise parameters for the satellite sensors or platform-induced distortions. Hence, a generic polynomial model is applied to register images to each other. We rely on polynomial distortion model since they are widely used as approximating functions in data analysis. A bivariate polynomial model relates the coordinates in the sensed image to those in the reference image. Let $x$ and $y$ be the scan and pixel coordinates of the sensed image, respectively. Also, $x_{ref}$ and $y_{ref}$ be the scan and pixel coordinates of the reference image, respectively. The generic polynomial model is given by,
\begin{align}
\begin{split}
x = \sum_{i=0}^{N} \sum_{j=0}^{N-i} a_{ij} x_{ref}^{i} y_{ref}^{j}, 
\\
y = \sum_{i=0}^{N} \sum_{j=0}^{N-i} b_{ij} x_{ref}^{i} y_{ref}^{j},
\end{split}
\end{align}
where, $a_{ij}$ and $b_{ij}$ are the coefficients for scan lines and pixels, respectively. The level of details in the data that can be approximated depends directly on the degree of polynomials, i.e., $N$. Root mean-squared (RMS) error is measured and compared up to $4^{th}$ degree of polynomials. From the GEO platform, the terrain variations of the earth do not appear to be significant and a $2^{nd}$ degree of polynomial approximation is sufficient to model the image data. Hence, a bivariate quadratic polynomial is used in our work. 

The quadratic polynomial equation is,
\begin{align}
\begin{split}
x &= a_{00} + a_{10} x_{ref} + a_{01} y_{ref} + a_{11} x_{ref} y_{ref} + a_{20} x_{ref}^{2} + a_{02} y_{ref}^{2},
\\ 
y &= b_{00} + b_{10} x_{ref} + b_{01} y_{ref} + b_{11} x_{ref} y_{ref} + b_{20} x_{ref}^{2} + b_{02} y_{ref}^{2},
\end{split}
\end{align}
where, the coefficients of quadratic polynomial contribute to total warp, i.e., shift, scale, shear, dependent scale, nonlinear scale, and nonlinear shear in $x$ and $y$. As the distortions in image data from satellites are random, we need to use least-square fit method. The matching pairs of GCPs have been established. Using these GCPs, the least-squares fit for the equation (6) determine the unknown coefficients.
%
This estimated transformation function (equation (6)) is now used to define a mapping from the reference image coordinates to the sensed image coordinates,
\begin{equation}
\left(x, y\right) = \mathbf{f}\left( x_{ref}, y_{ref}\right), \forall x_{ref},y_{ref}.
\end{equation}

The mapping functions constructed are used to transform the sensed image and thus to register the image. Since, the transformed coordinates will not be integer, in general, a new pixel must be estimated between existing pixels by an appropriate interpolation process known as {\itshape {resampling}}. Basically, it is the interpolation technique of generating an image on a system of coordinates, taking the input image from a different set of coordinates. The corrected output image is then created by filling each location  of a new, initially empty array, with the pixel calculated by transformed coordinates in the sensed image.	

There are various resampling (interpolation) methods available for digital imagery\cite{4342630}. Selection of an appropriate resampling technique can be treated as application dependent criterion. nearest neighbour (NN) resampling is used in our application. NN is the fastest scheme compared to other resampling techniques discussed. The other techniques put more computational load on the program and consequently slows down the registration process, besides tampering with the actual gray values. On the other hand, NN resampling introduces blocking artifacts in the processed image, because of its ``round-off" property, i.e., geometrical discontinuities of the order of $\pm \frac{1}{2}$ pixels. Nevertheless, NN resampling does not introduce new gray values. Hence, the image statistical distribution remains almost unchanged. So, the processed image can be used for further applications, where radiometric accuracy is also of importance, such as in our meteorological purposes.

{\itshape A flowchart of software implementation:}
A simplified flowchart of the complete software implementation is given in Fig. 5. As shown in the flowchart, the sensed images are first read into two-dimensional arrays. Match points are identified either operating on the image data or after extracting edges in both the images. Moving window algorithm has been implemented to automate the match points identification in the sensed image. Then, the sensed image is warped, by the transformation function (equation (6)), so as to geometrically fit into the reference image coordinates. The sensed image is then resampled to the transformed coordinate system by using the NN as discussed above. Thus, a sensed image is registered with respect to a reference image.
\begin{figure*}[h!]
	\centering
		\includegraphics[width=\textwidth]{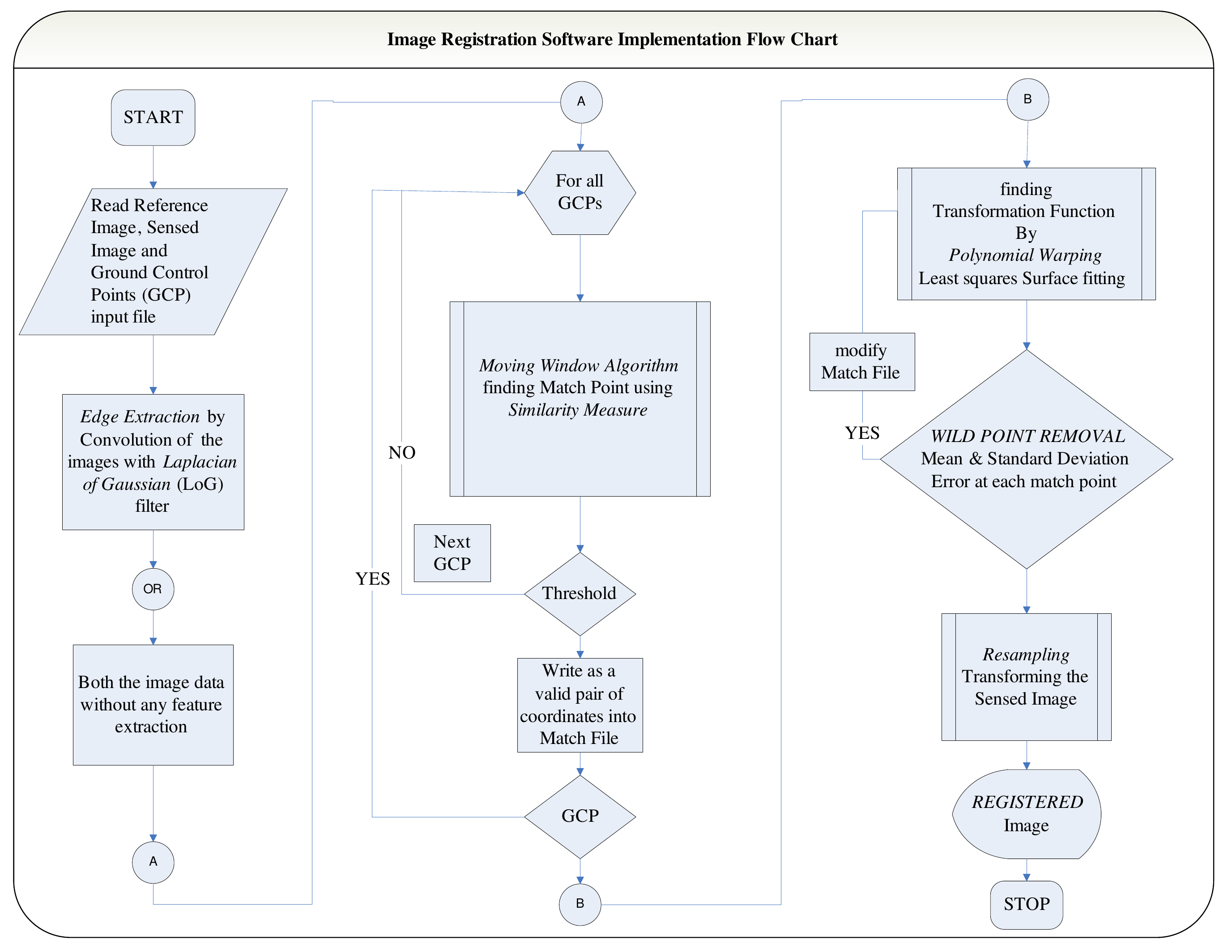}
	\caption{A simplified flow chart for the complete registration process.}
	\label{fig:data_flow_dia}
\end{figure*}

\section{Results}
This section discusses the data registration of images from KALPANA-1 and INSAT-3A. The methods have been rigorously tested on  different images including multiband and multitemporal data. We reproduce few important results in this section. The development of the algorithms are done in C-programming running on a 32-bit Linux operating system. All the algorithms are run on a server machine with Intel\text{\textregistered} Xeon$^{\rm TM}$ Dual CPU at 3.20 GHz with 4 GB of RAM. Image Display Software developed by SAC (ISRO) is used for displaying the images. 

\subsection{Results using INSAT-3A CCD Data}
This subsection demonstrates the results of INSAT-3A CCD data registration using the proposed approach. A case for multitemporal image registration is shown in Fig. 6 for a CCD image. Fig. 6 (a) and (b) show a synoptic view of a reference and a sensed image, respectively, in the display panels. The sensed image is to be registered with respect to the reference image. Non-linear distortions and corresponding registrations at different points in the sensed image are shown in Fig. 7 (a) to (c). 

Fig. 7 (a - left panel) shows a distortion in the sensed image at current cursor position, with respect to the same point in the reference image. As can be seen, the sensed point 1 is shifted 31 scan lines down side and 08 pixels right side, with the respect to the reference point 1. After applying our registration algorithm, the geometrically shifted sensed point 1, shown in Fig. 7 (a - left panel), is aligned with respect to the reference point 1, as shown in Fig. 7 (a - right panel), i.e., Scan: 2641, Pixel: 3031. Besides, the pixel brightness value of the sensed image is preserved at the registered point 1, i.e., 0048, referring to Fig. 7 (a). Fig. 7 (b) and (c) demonstrate registration at two more points. It should be noted that geometric distortions observed at Fig. 7 (b) and (c) are non-linear when compared to the point in Fig. 7 (a). Similar results have obtained for all the GCPs in the sensed image and finally the sensed image is registered with respect to the reference image. 

\begin{figure*}[h!]
	\centering
		\includegraphics[width=0.8\textwidth]{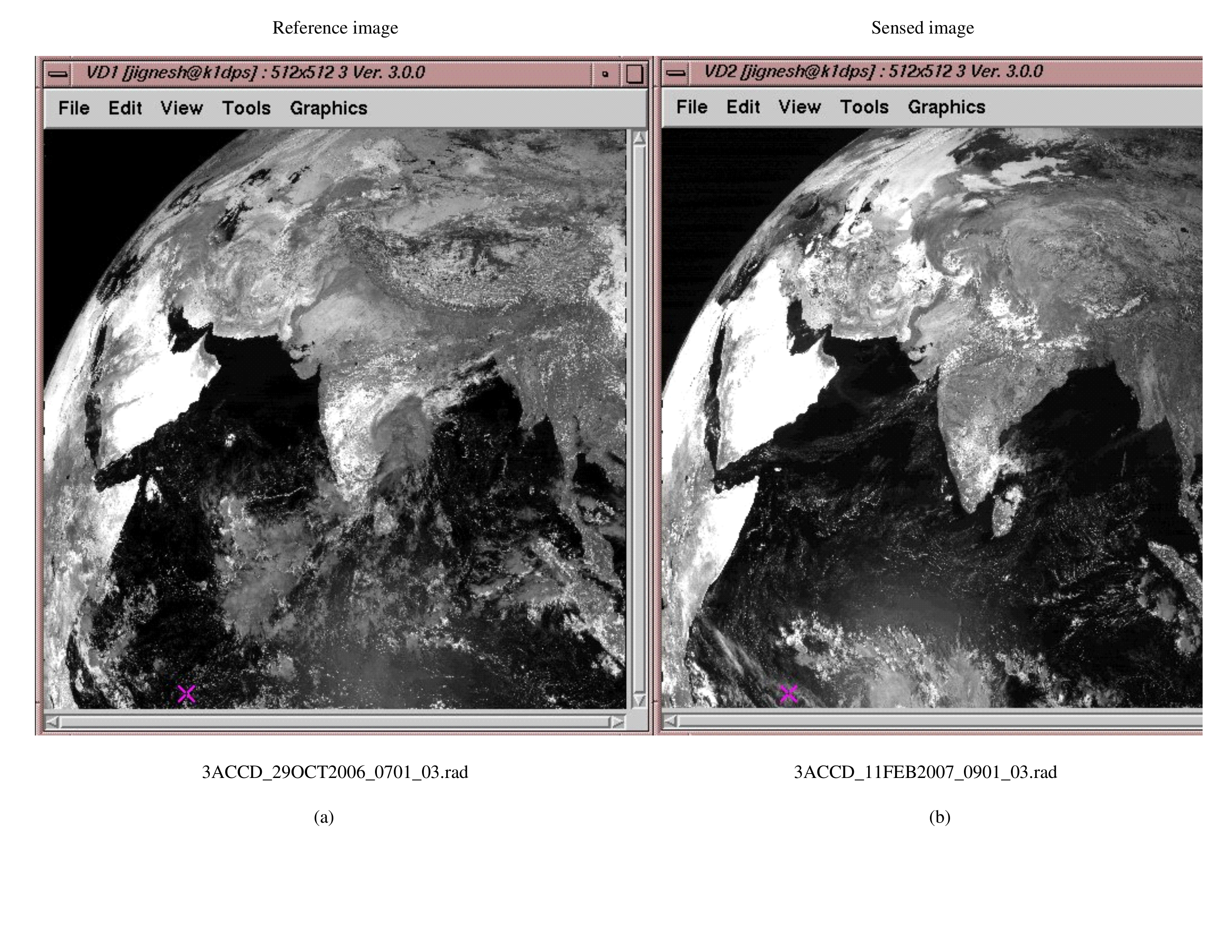}
	\caption{INSAT-3A CCD data registration. (a) reference image, and (b) sensed image to be registered with (a). }
	\label{fig:CCD_reg_input}
\end{figure*}

\begin{figure*}[h!]
	\centering
		\includegraphics[width=0.8\textwidth]{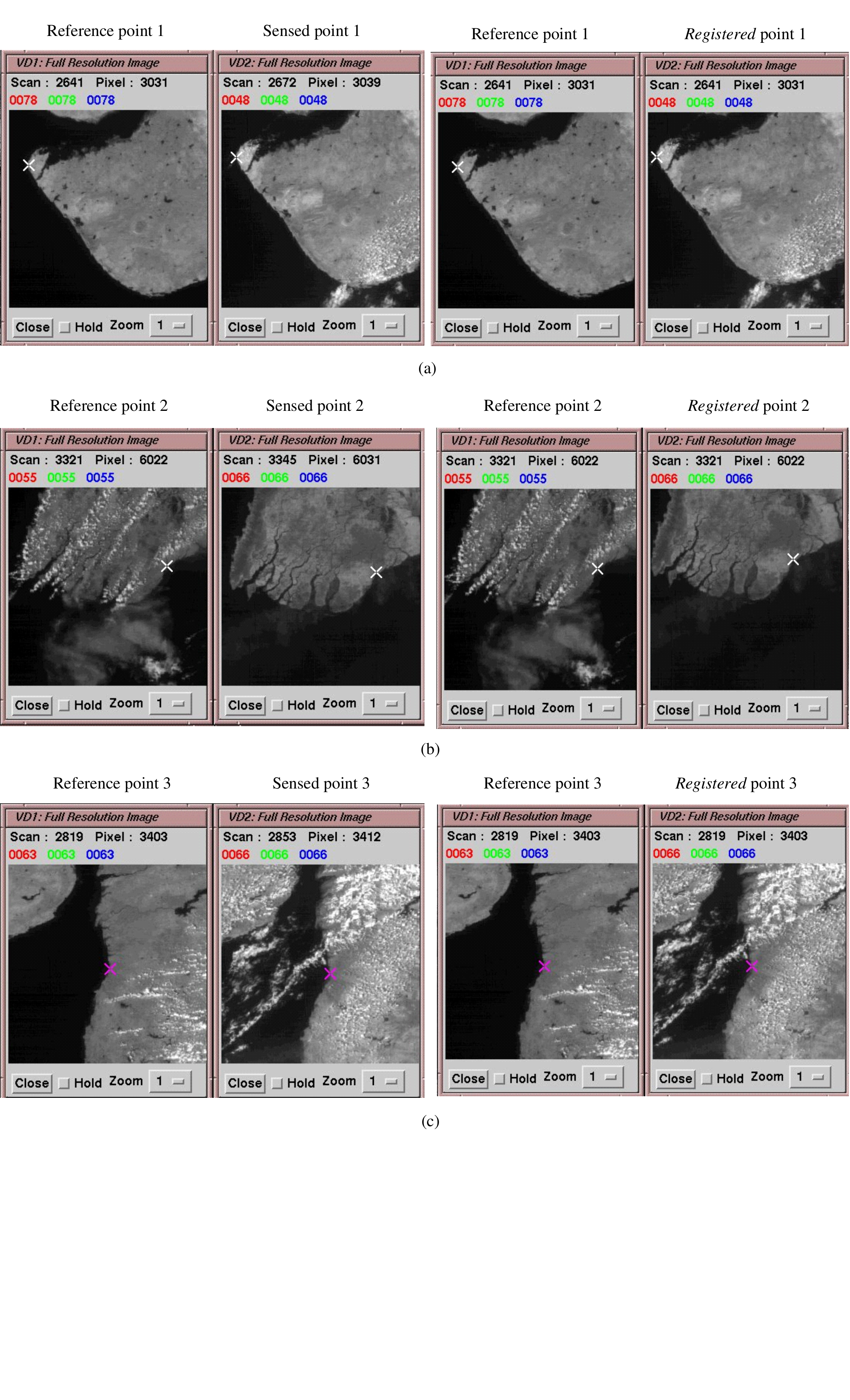}
	\caption{INSAT-3A CCD data registration at diiferent points over the imagery: Three different points (a), (b), and (c) are registered with their respective GCPs marked on corresponding reference points.}
	\label{fig:CCD_reg_points}
\end{figure*}

\subsection{Reference boundaries generation for INSAT-3A CCD Data}
After an image is registered with a reference image, boundaries are to be overlaid over the registered image. As the boundary for CCD images was not available, it has been generated from the VHRR boundary image data from the same satellite, i.e., INSAT-3A. This subsection discusses about the reference boundaries generation and shows the related results on the INSAT-3A CCD data. VHRR reference boundaries (full globe product) is generated from the Shape file provided by the IMD, New Delhi.

A $4000\times4000$ size image is first cut from the VHRR full globe product, with the same view as of CCD images. The cut image has been treated as a sensed image for the warping. The reference image is made of $8000\times8500$, by padding zeros all around the image area. Points on the Indian coast and surrounding were identified manually and used the warp (equation (6)) with respect to the reference image $8000\times8500$ in order to generate the boundary image. Fig. 8 shows a synoptic view of the reference CCD image and its generated boundary image.
\begin{figure*}
	\centering
		\includegraphics[width=\textwidth]{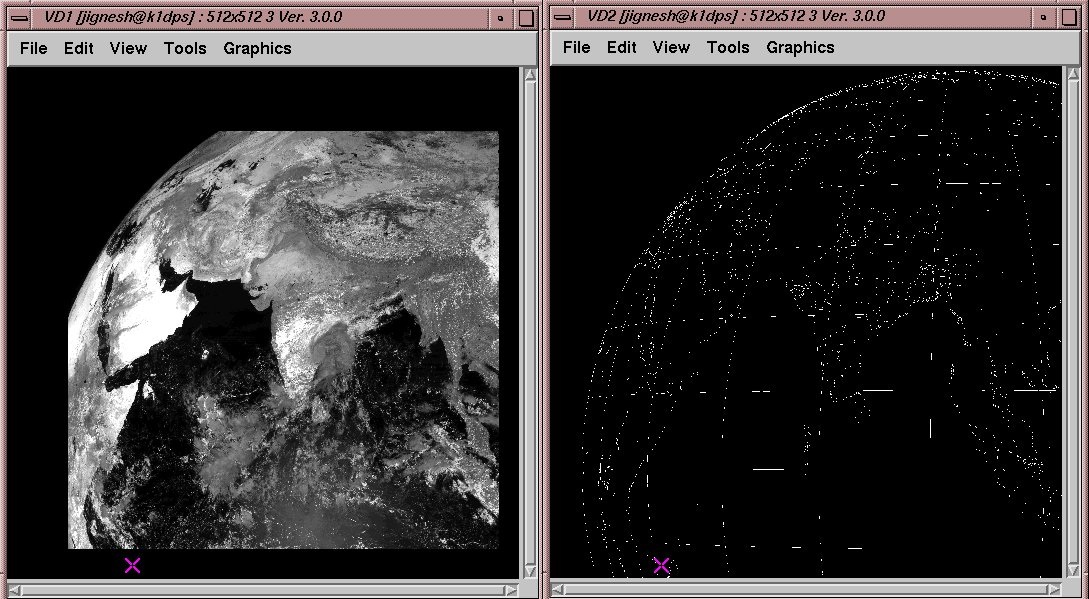}
	\caption{Synoptic view of the CCD reference image and its generated boundaries using proposed warping algorithm.}
	\label{fig:CCD_reg_points}
\end{figure*}

The generated boundaries are then overlaid over the reference image. Fig. 9 shows the CCD boundary overlaid over the reference image and Fig. 10 shows two full resolution snapshots of the boundaries fitting on the image.

\begin{figure*}
	\centering
		\includegraphics[width=\textwidth]{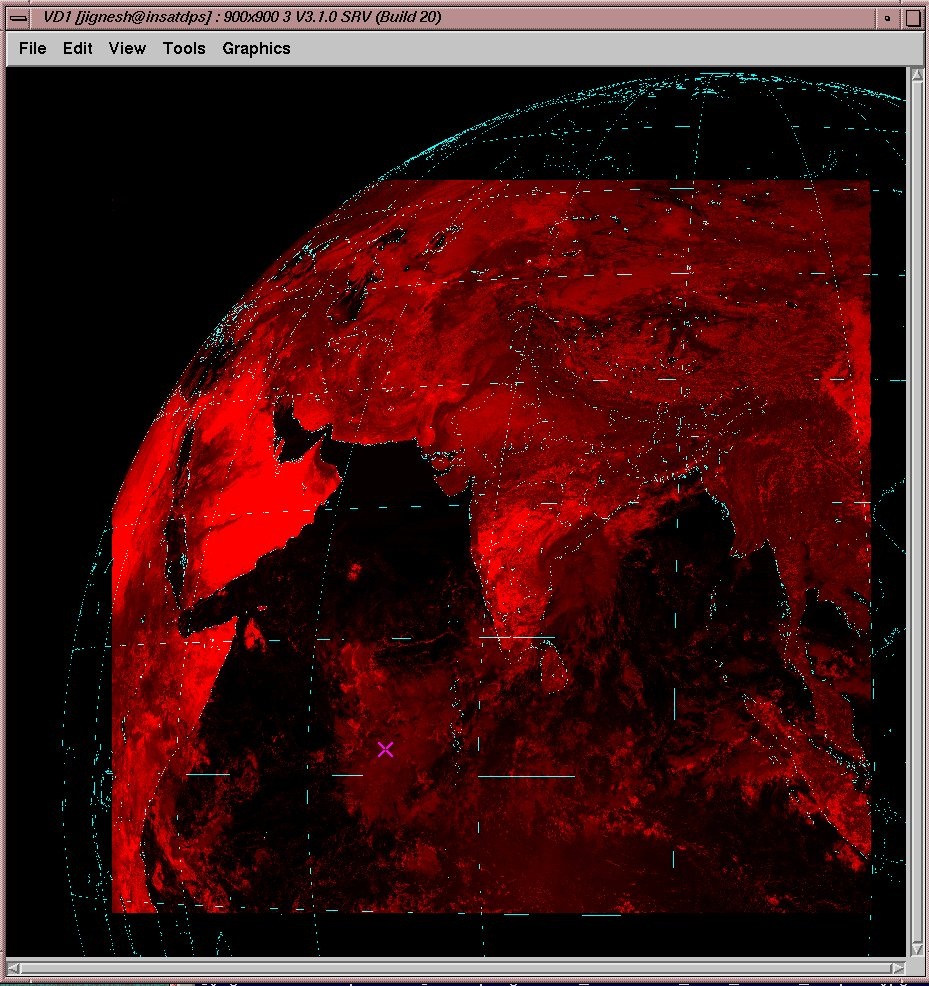}
	\caption{CCD boundaries overlaid on the reference image referring to Fig. 8.}
	\label{fig:CCD_reg_points}
\end{figure*}

\begin{figure*}
	\centering
		\includegraphics[width=\textwidth]{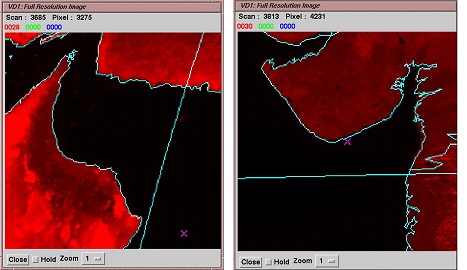}
	\caption{Full resolution snapshots of the CCD image boundary fitting referring to Fig. 9.}
	\label{fig:CCD_reg_points}
\end{figure*}

\subsection{Results using KALPANA-1 VHRR Data}
In this subsection, we show the results of data registration for KALAPANA-1 VHRR imagery. Fig. 11 shows a synoptic view of a reference and a sensed image in display panels. The sensed image (right panel in Fig. 11) is to be registered with respect to the reference image (left panel in Fig. 11). A distortion and corresponding registration in the sensed image is shown in Fig. 12. Finally, the boundaries are overlaid on the registered image as shown in Fig. 13.
\begin{figure*}
	\centering
		\includegraphics[width=\textwidth]{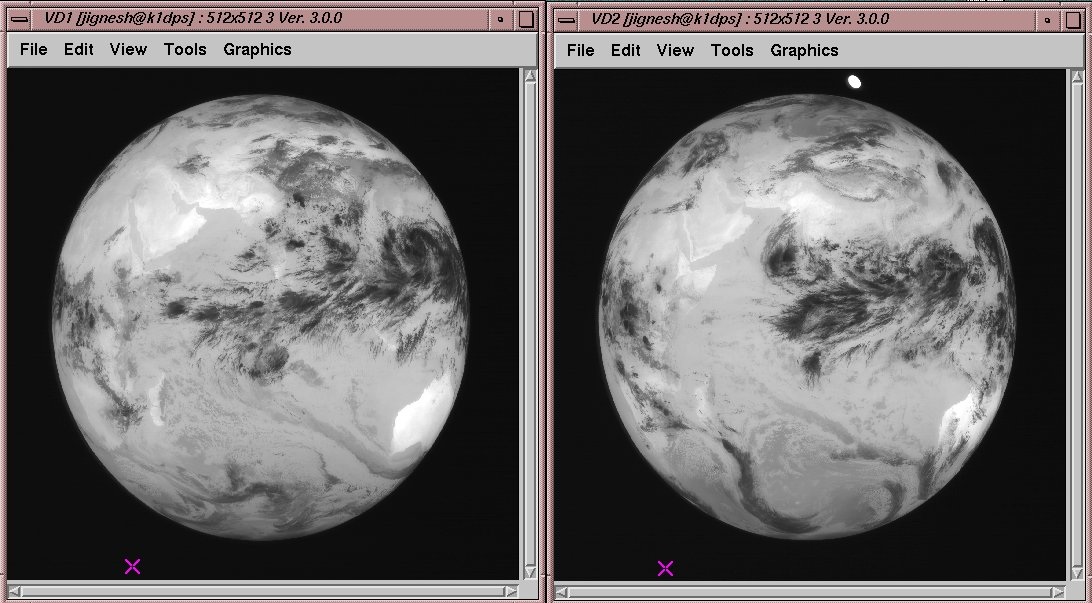}
	\caption{Synoptic view of the reference image (left panel - K1VHR\_29SEP2007\_0600\_IR.rad) and the sensed image (right panel - K1VHR\_25OCT2007\_0600\_IR.rad) from KALPANA-1 VHRR for data registration.}
	\label{fig:CCD_reg_points}
\end{figure*}
\begin{figure*}
	\centering
		\includegraphics[width=\textwidth]{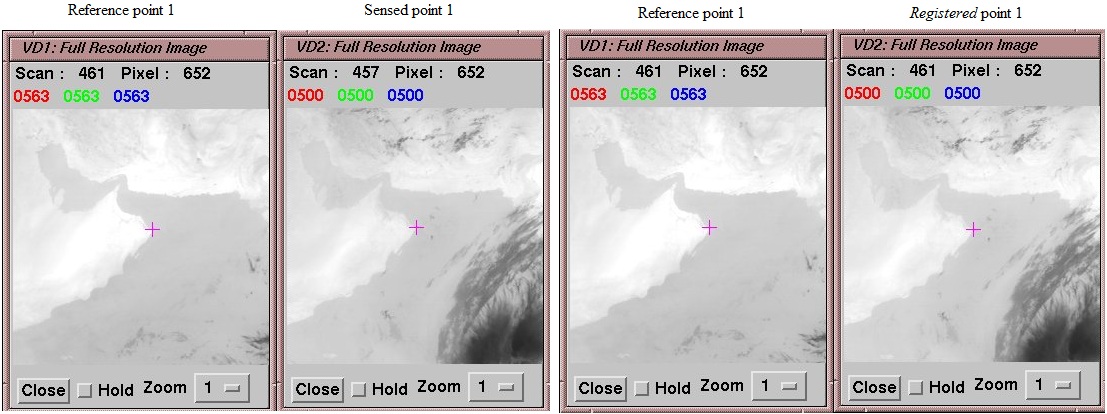}
	\caption{KALPANA-1 VHRR data registration at a point.}
	\label{fig:CCD_reg_points}
\end{figure*}
\begin{figure*}
	\centering
		\includegraphics[width=\textwidth]{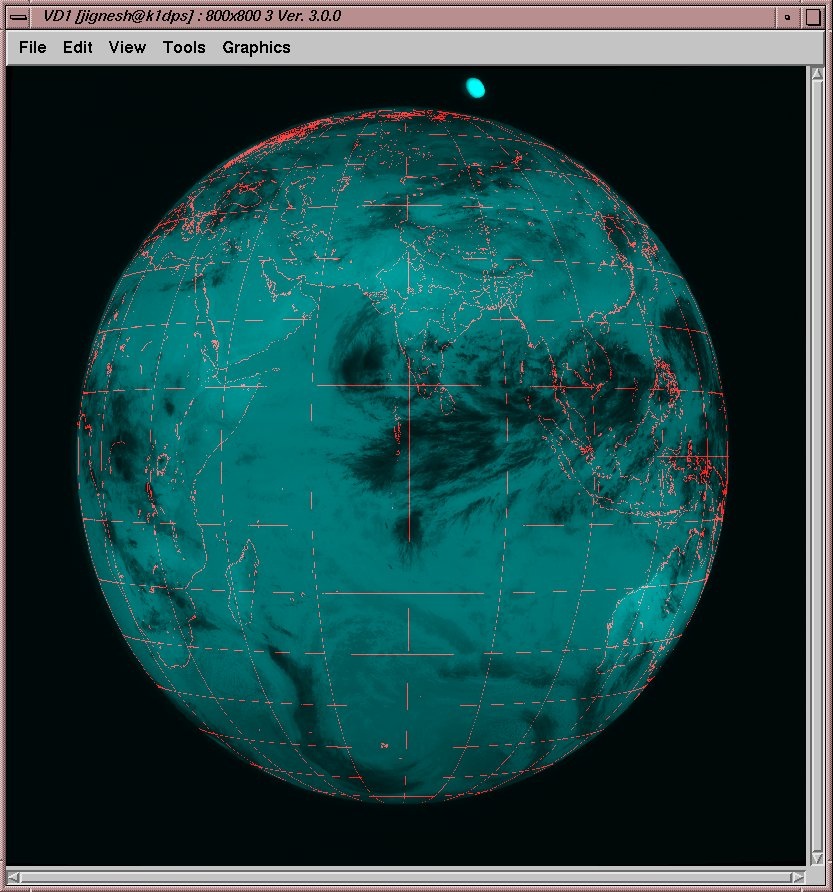}
	\caption{VHRR boundaries overlaid on the registered image from KALPANA-1 payload.}
	\label{fig:CCD_reg_points}
\end{figure*}

\subsection{Performance comparison of the similarity measures}
In this subsection, the similarity measures are compared for both VHRR and CCD images. As discussed in Section IV, the moving window algorithm has been implemented to automate the match points identification. To this end, the size of the search window should be selected so as not to miss any GCP for corresponding match identification in a sensed image. Besides, the sizes of target window and search window should not be too large and too small as well. By looking at the amount of deviations in the VHRR images, optimal sizes of the windows have been selected. The size of target window is $11\times11$ and a search window of $31\times31$ has been utilized. Table 1 shows the performance comparisons of various similarity measures for VHRR images.
\begin{table*}[ht]
\caption{Performance comparison of similarity measures in data registration}
\centering
\begin{tabular}{c | c |c| c| c|c}
\hline\hline
Data & Similarity measures & Total input GCPs & Match points found & RMSE\cite{Wang_Bovik_2009} (scan,pixel) & Execution time \\
\hline
VHRR & MI & 29 & 27 & (2.92, 3.13) & 8.5 minutes \\
VHRR& CRA & 29 & 28 & (2.98, 2.65) & 2.5 minutes \\
VHRR& combination of NCC and MSD & 29 & 29 & (2.5, 2.1) & 0.6 seconds\\
CCD & combination of NCC and MSD & 26 & 23 & (8.8, 6.7) & 12.5 seconds\\
\hline
\hline
\end{tabular}
\label{table: simulate data}
\end{table*}

In case of CCD images, there are more violent deviations when compared to the VHRR data. Hence, sizes of the windows have to be increased. Hence in the CCD case, search window of $101\times101$ and target window of $21\times21$ have been selected. However, this leads to higher computational time for automatic match points identification, specially, in cases of MI and CRA. Table 1 shows the performance of combination of NCC and MSD as a measure for the CCD image registration. Note that we have avoided MI and CRA in the comparison for CCD data. 

\subsection{Multiband data registration for VHRR and CCD images}
Multiband registration, for the images acquired at a same time of a day, has been achieved in the similar way as for the multitemporal registration. An image is taken as reference and all other images, at a same time from different bands for a day, are registered with respect to the reference one. The same algorithm has been used for the multiband image registration. We have not reproduced the results of the same due to space constraint.

\section{Conclusions}
Registration is one of the fundamental steps while deriving software products from the acquired data. We have developed a generalized software for data registration of both CCD and VHRR sensors. The algorithm is fast, accurate as well as adaptive enough for updation in future. In addition the methodology can also be adopted to other modalities of remote sensing data.

\section*{Acknowledgment}
Authors are grateful to the Department of Space, Government of India, for allowing to carry out this work at SAC (ISRO) and making available the VHRR/CCD sensors datasets from KALPANA-1 and INSAT-3A payloads. We are thankful to Mr. Tapan Misra, Director, SAC (ISRO), Ahmedabad, for allowing us to publish the work. The authors thank to Dr. R. Ramakrishanan, Group Head (DPSG) for his encouragement and support. Special thanks to Mr. Sajid Mohammad for his valuable contribution in the preparation of this paper.

\bibliographystyle{IEEEtran}   
\bibliography{myref}   

%

\end{document}